\documentclass[runningheads]{llncs}

\usepackage{eccv}

\usepackage{eccvabbrv}

\usepackage{graphicx}
\usepackage{booktabs}

\usepackage[accsupp]{axessibility}  

\usepackage{hyperref}

\usepackage{orcidlink}

\begin{document}

\title{Distractors-Immune Representation Learning with Cross-modal Contrastive Regularization \\ for Change Captioning} 

\titlerunning{DIRL with CCR for Change Captioning}

\author{Yunbin Tu\inst{1} \and
Liang Li\inst{2*} \and
Li Su\inst{1*} \and
Chenggang Yan\inst{3} \and
Qingming Huang\inst{1}}

\authorrunning{Y.Tu et al.}

\institute{University of Chinese Academy of Sciences, Beijing, China 
\email{tuyunbin22@mails.ucas.ac.cn,} \email{\{suli,qmhuang\}@ucas.ac.cn}
\and
Key Laboratory of AI Safety of CAS, Institute of Computing Technology, Chinese Academy of Sciences (CAS), Beijing, China
\email{liang.li@ict.ac.cn}\\
\and
Hangzhou Dianzi University (HDU) \& Lishui Institute of HDU, China}
\maketitle
\renewcommand{\thefootnote}{\fnsymbol{footnote}}
\footnotetext[1]{Corresponding authors}

\begin{abstract}
 Change captioning aims to succinctly describe the semantic change  between a pair of similar images, while being immune to distractors (illumination and viewpoint changes). Under these distractors, unchanged objects often appear pseudo changes about location and scale, and certain objects might overlap others, resulting in perturbational and  discrimination-degraded features between two images. However, most existing methods directly capture the difference between them, which risk obtaining error-prone difference features. In this paper, we propose a distractors-immune representation learning  network that correlates the corresponding channels of two image representations and decorrelates different ones in a self-supervised manner, thus attaining a pair of stable image representations under distractors. Then, the model can better interact them to capture the reliable difference features for caption generation. To yield words based on the most related difference features, we further design a cross-modal contrastive regularization, which regularizes the cross-modal alignment by maximizing the contrastive alignment between the attended difference features and   generated words. Extensive experiments show that our method outperforms the state-of-the-art methods on four public datasets. The code is available at \url{https://github.com/tuyunbin/DIRL}.
\end{abstract}

\section{Introduction}
\label{sec:intro}

\begin{figure}[t]
  \centering
   \includegraphics[width=1\linewidth]{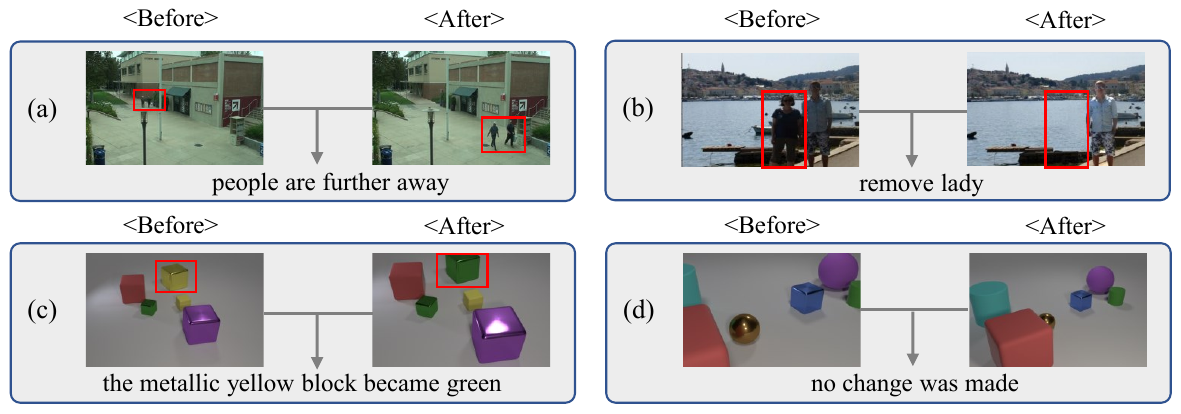}

   \caption{The examples of change captioning under different scenarios. The first and second cases show that object moving and dropping, respectively. The third one shows the color change under distractors (viewpoint and illumination changes), where the real change is overwhelmed by pseudo changes. The last one shows with only distractors. Changed objects are shown in  red boxes.}
   \label{fig1}
\end{figure}

Recently, researchers have witnessed the great success toward vision-language understanding and generation \cite{yang2022paraphrasing,liu2023entity,tang2024context,xiao2024r}. As an emerging task, change captioning \cite{sun2024stvchrono,yue2024multi,tu2024context} is to describe what has semantically changed between two similar images in natural language (Fig. \ref{fig1} (a)-(d)). This task has a wide range of practical applications, such as generating reports for surveillance area changes \cite{hoxha2022change} and pathological changes between medical images \cite{li2023dynamic}, as well as providing visually-impaired users with the explanations of image editing effects \cite{tan2019expressing}. 

This task also poses a formidable challenge: models should be powerful enough to not only  understand the contents of two images, but also describe the semantic change between them, while being immune to distractors (viewpoint and illumination changes). In a dynamic environment, two images of the same scene are usually obtained in the presence of distractors. In this situation, unchanged objects between two images often appear obviously pseudo changes about the scale and location (Fig.  \ref{fig1} (c) (d)), where the features of same objects within the image pair might be perturbational. Especially, as a viewpoint drastically changes, certain objects might partially overlap the others, \emph{e.g.,} the brown ball is partially occluded by the red block in the ``after'' image of Fig.  \ref{fig1} (d). Accordingly, the feature discrimination might be weakened under this viewpoint. In short, there are perturbational and discrimination-degraded features between two image representations under distractors. 

Most previous methods directly subtract between such two image representations \cite{park2019robust,hosseinzadeh2021image,tu2021semantic,sun2022bidirectional} or compute their feature similarity \cite{shi2020finding,qiu2021describing,yao2022image,yue2023i3n}, which risk capturing error-prone difference features in between.  The latest works SCORER \cite{tu2023self} and SMART \cite{tu2024smart} overcome the distractors by implementing contrastive learning between similar/dissimilar image pairs. By maximizing the alignment of similar ones, both methods make the features of unchanged objects non-perturbational under distractors. However, both methods disregard the feature discrimination-degraded problem, leaving the influence of distractors upon a pair of image representations remained.

In this paper, we propose a \textbf{D}istractors-\textbf{I}mmune \textbf{R}epresentation \textbf{L}earning (DIRL)  network to attain a pair of stable image representations that are non-perturbational and discriminative under distractors, for robust change captioning. Concretely, given two raw image representations, DIRL first  computes a channel correlation matrix between them.  Next, DIRL implements cross-channel decorrelation to optimize this  matrix to be close to the identity matrix, \emph{i.e.,} making the corresponding channels of two image representations  have similar semantics, while enforcing  different channels to be independent. In this self-supervised fashion, DIRL not only alleviates feature perturbation between two images, but also enhances feature discrimination for each image. As such, the model can sufficiently interact the two stable image representations to infer their reliable difference features, which are then translated into a linguistic sentence via a transformer decoder.

During sentence generation,  as the changed object  typically occurs  in   a local region  with weak feature, traditional attention mechanisms used in mainstream methods \cite{kim2021agnostic,qiu2021describing,tu2023viewpoint}  may lead to less satisfactory cross-modal alignment. To facilitate correct alignment, we further design a \textbf{C}ross-modal \textbf{C}ontrastive \textbf{R}egularization  (CCR), which is built upon the cross-attention module of  transformer decoder. When the cross-attention module yields the attended difference features, CCR further regularizes them by maximizing the contrastive alignment between them and  generated words, which helps the decoder generate the change caption  based on the most related difference features. 
 
 \textbf{Our key contributions are}: \textbf{(1)} We propose the novel DIRL network to learn a pair of stable image representations under distractors by correlating their corresponding channels and decorrelating different channels. Based on the two distractors-immune representations, the model is able to learn the robust difference features between them for caption generation. \textbf{(2)} We design the CCR to regularize the cross-modal alignment  by maximizing the contrastive alignment between the features of attended difference and generated words. This helps the decoder generate words based on the most related difference features. \textbf{(3)} Our method performs favourably against state-of-the-art methods on four public datasets with different change scenarios.

\section{Related Work}


Change captioning is a new task in the vision-and-language area  \cite{cho2022fine,tu20222,li2022long,tu2023relation,islam2024video,chen2024prompt}. The pioneer work \cite{jhamtani2018learning} releases the dataset about changes in surveillance scenarios with potential illumination changes. Afterwards, Tan \emph{et al.} \cite{tan2019expressing} collect a dataset about image editing scenes. Both works  \cite{park2019robust,kim2021agnostic} develop two datasets for simulating distractors (viewpoint and illumination changes), where the unchanged objects' scale and location would illustrate obviously pseudo changes.

To learn the difference features under distractors, prior works \cite{park2019robust,hosseinzadeh2021image,liu2022remote} directly subtract between two image representations, which are hard to generalize to unaligned image pairs. 
SGCC \cite{liao2021scene}  introduces 3D information of object depths to overcome viewpoint changes. Recent works \cite{shi2020finding,kim2021agnostic,yue2023i3n,tu2023viewpoint} first measure the feature similarity to summarize the shared features and then remove them to obtain  difference features. On the other hand, the works \cite{qiu2021describing,yao2022image,guo2022clip4idc} correlate the similar features of two images to implicitly deduce the difference features. However, the above methods do not well address the influence of distractors upon image representations. 
The latest works SCORER \cite{tu2023self} and SMART \cite{tu2024smart} try contrastive  learning between similar/dissimilar image pairs, which model the alignment between similar ones  to handle distractors. There are two differences between both works and our DIRL. First, SCORER and SMART capture relations  between all paired/unpaired images in a batch, whereas  DIRL computes the correlation between the feature channels of each image pair. Second, SCORER and SMART only make  the features between two images non-perturbational under distractors, while DIRL additionally considers   enhancing feature discrimination for each image to better recognize  semantic changes.


During caption generation, since the changed object typically appears in a local region  with weak feature, it is difficult to learn reliable alignment via current attention modules in prevalent methods. To address this issue,  SMART \cite{tu2024smart} employs part-of-speech to guide the  decoder to dynamically use visual information. These works \cite{hosseinzadeh2021image,kim2021agnostic,tu2023self} propose a cycle consistence module to enforce correct cross-modal alignment. The other works \cite{yao2022image,guo2022clip4idc,black2024vixen} design a pre-training and fine-tuning strategy to strengthen the alignment. 
In this paper, we regularize cross-modal alignment by proposing CCR, which maximizes the contrastive alignment between the generated words and  attended difference features. We find that two works in single-image captioning \cite{cho2022fine,yang2022paraphrasing} tried the similar idea.  Both works model the coarse-grained cross-modal contrastive regularization between the input image and generated caption. Instead,  our CCR reuses the attended difference features; establishes the fine-grained contrastive regularization between them and generated words during decoding. In this way, the model can use the most relevant difference features to predict next words.

\begin{figure*}[t]
\centering
\includegraphics[width=1\textwidth]{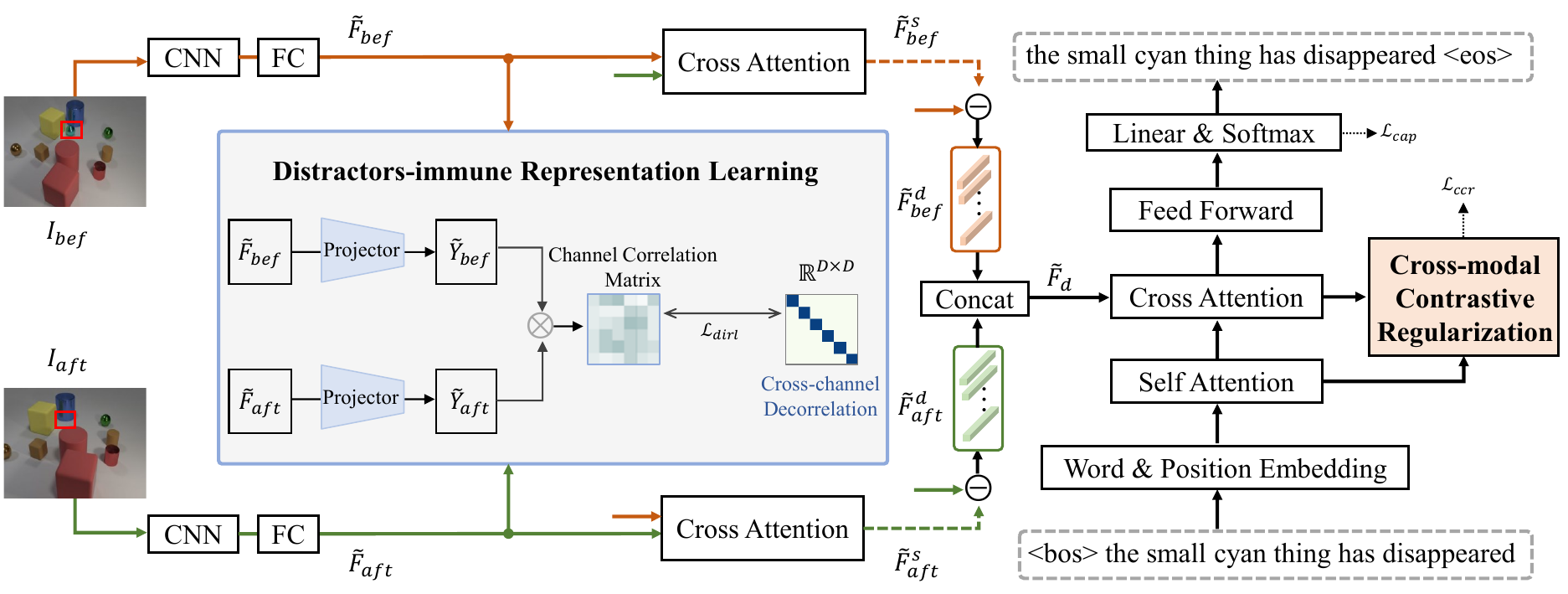} 
  \caption{The framework of our method, where the core blocks are \textbf{distractors-immune representation learning} network and \textbf{cross-modal contrastive  regularization}. FC and Concat are short for the fully-connect layer and concatenation operation.  }
\label{fig2}
\end{figure*}

\section{Methodology}

As shown in Fig. \ref{fig2}, the  overall framework of our method includes following parts. First, a pre-trained CNN  model extracts the raw representations of an image pair, which are then transformed into a low-dimensional space. Next, the two image representations are fed into the proposed DIRL to make them non-perturbational and discriminative  under distractors. Subsequently, a cross-attention module interacts the two stable  representations to mine the shared features, which are removed to learn the  difference features for decoding into a sentence. During sentence generation, the designed CCR regularizes the cross-modal alignment, so as to help the decoder generate the sentence based on the most related difference features.

\subsection{Image Pair Feature Extraction}
Given a pair of images ``before'' $I_{bef}$ and ``after'' $I_{aft}$, we first use a pre-trained CNN  to extract their feature representations. They are denoted as $F_{bef}$ and $F_{aft}$, where ${F}_{o} \in \mathbb{R}^{C \times H \times W}$. $C$, $H$, $W$ indicate the number of channels, height, and width. Then, we  project them into a low-dimensional space of $\mathbb{R}^{D}$ by a 2D-convolution:
\begin{equation}
\tilde {F}_{o}=\text{conv}_2 ({F}_{o})  +\text{pos}({F}_{o}),\\
\end{equation}
where $o \in (bef,aft)$. pos is the learnable position encoding function.

\subsection{Distractors-immune Representation Learning}
After obtaining the two position-embedded image representations, we propose a DIRL network to make them non-perturbational and discriminative under distractors in a self-supervised manner.
Concretely, we first project $\tilde F_{bef}$ and $\tilde F_{aft}$ into a common embedding space with shared parameters:
\begin{equation}
\tilde Y_o= \text{MLP} \left(\tilde F_o\right), o \in(b e f, a f t),
\end{equation}
where MLP is a two-layer multi-layer perceptron with the ReLU activation function in between. Then, we compute a channel correlation matrix between $\tilde Y_{bef}$ and $\tilde Y_{aft}$ along the batch dimension:

\begin{equation}
\mathcal{C}_{i j} = \frac{\sum_b \tilde y_{b, i}^{bef} \tilde y_{b, j}^{aft}}{\sqrt{\sum_b\left(\tilde y_{b, i}^{bef}\right)^2} \sqrt{\sum_b\left(\tilde y_{b, j}^{aft}\right)^2}},
\end{equation}
where $b$ indexes batch samples and $i, j$ index the channel dimension of features. $\mathcal{C} \in \mathbb{R}^{D\times D}$ is a channel correlation matrix. 
Inspired by the recent self-supervised learning methods \cite{zbontar2021barlow,wang2022disentangled}, we perform cross-channel decorrelation to enforce this matrix to be close to the identity matrix, which is implemented by using the $\ell_2$-norm minimization:
\begin{equation}
\label{alpha}
\mathcal{L}_{dirl} = \sum_i\left(1-\mathcal{C}_{i i}\right)^2+\alpha  \sum_i \sum_{j \neq i} \mathcal{C}_{i j}^2.
\end{equation}
The first term  equates the diagonal of $\mathcal{C}$ to one, \textit{i.e.,} the corresponding channels of two image representations will be correlated and thus have similar semantics under distractors.
 The second term equates the off diagonal of $\mathcal{C}$ to zero, \textit{i.e.,} the different channels will be decorrelated. This enhances the discrimination of each image representation. $\alpha$ is a trade-off parameter to balance the importance between two terms, which is discussed in the supplementary material. 

\subsection{Difference Representation Learning}
The difference features are modeled based on the two distractors-immune image representations. As shown in Fig. \ref{fig1}, most objects are identical between two images. Hence,  it is natural to extract their shared features to infer the difference features. For a powerful model, it should capture all potential changes w.r.t. both images. To this end, we first  learn the difference features for each image and then combine them to construct the difference features between two images. 

Specifically,  we first compute the shared features on each image by the multi-head cross-attention (MHCA) mechanism \cite{vaswani2017attention}:
\begin{equation}
\begin{gathered}
\tilde{F}_{b e f}^s=\operatorname{MHCA}\left(\tilde{F}_{b e f}, \tilde{F}_{a f t}, \tilde{F}_{a f t}\right), \\
\tilde{F}_{a f t}^s=\operatorname{MHCA}\left(\tilde{F}_{a f t}, \tilde{F}_{b e f}, \tilde{F}_{b e f}\right).
\end{gathered}
\end{equation}
Subsequently, we subtract each from the corresponding image representation to compute the difference features on each image, respectively:
\begin{equation}
\begin{gathered}
\tilde{F}_{b e f}^d=\tilde{F}_{b e f} - \tilde{F}_{b e f}^s,\\
\tilde{F}_{a f t}^d=\tilde{F}_{aft} - \tilde{F}_{aft}^s.
\end{gathered}
\end{equation}
Both $\tilde F_{bef}^d$ and $\tilde F_{aft}^d$\ are then concatenated as the  omni-representation of difference features between two images, which is implemented by a fully-connected layer with the ReLU activation function:
\begin{equation}
\tilde{F}_{d}=\operatorname{ReLU}\left(\left[\tilde{F}_{bef}^{d} ; \tilde{F}_{aft}^{d}\right] W_{c} + b_c \right ),
\end{equation}
where [;] is a concatenation operation. 

\subsection{Caption Generation}
After learning $ \tilde{F}_{d} \in \mathbb{R}^{HW \times D}$, we use a standard transformer decoder \cite{vaswani2017attention} to translate it into a sentence. First, we obtain the embedding features of $m$ words (ground-truth words during training, predicted words during inference). Then, we use the multi-head self-attention (MHSA) layer  to model  relationships among these word embedding features $ E[W]=\{E[w_1],... ,E[w_{m}]\}$:
\begin{equation}
 {\hat E[W]}=\text { MHSA }(E[W],E[W],E[W]).
\end{equation}
Subsequently, we use the relation-embedded word features $ \hat E[W]$ to attend to the related difference features from $ \tilde{F}_{d} $ based on the MHCA layer:
\begin{equation}
\label{mhca}
\hat V=\text { MHCA }({E[\hat W]},\tilde{F}_{d}, \tilde{F}_{d}).
\end{equation}
Next, the $\hat V \in \mathbb{R}^{HW \times D}$ is passed to a feed-forward network (FFN) to obtain the enhanced difference features:
\begin{equation}
\hat V' =\text { LayerNorm}(\hat V+\operatorname {FFN}(\hat V)).
\end{equation}
Finally, the probability distributions of target words are calculated via a single hidden layer:
\begin{equation}
\label{word}
W=\operatorname{Softmax}\left(\hat V' W_{h}+{b}_{h}\right),
\end{equation}
where $W_{h}\in \mathbb{R}^{D \times \nu }$ and $b_{h} \in \mathbb{R}^{\nu }$ are the learnable parameters. $\nu $ is the dimension of vocabulary size.

\subsection{Cross-modal Contrastive Regularization}
It is often the case that the changed object   appears in a local area  with weak feature, so it is difficult for the cross-attention module to directly align the captured difference features with word features. 
To this end, we devise the CCR  to regularize the cross-modal alignment, which reuses the attented difference features $\hat V$ computed in Eq. (\ref{mhca}) and maximizes  the contrastive alignment between them and generated words ${\hat E[W]}$. Specifically, we first compute the global representation for  ${\hat E[W]}$ and $\hat V$ via mean-pooling operation, respectively:
\begin{equation}
\begin{gathered}
\tilde E[W]=\frac{1}{m} \hat E[W], \\
\tilde V=\frac{1}{HW} \hat V,
\end{gathered}
\end{equation}
where $m$ and $HW$ are the length of ${\hat E[W]}$ and $\hat V$. Then, given a training batch, we sample $B$ pairs of global representations of generated words and attended difference features. For $k$-th global word representation  $\tilde E[W]_k$,  $k$-th global difference representation $\tilde V_k$ is its positive, while the others will be the negatives in this batch. The similarity between two global representations is computed by dot-product:
\begin{equation}
\text{sim}\left(\tilde E[W], \tilde V\right) =\tilde E[W] \cdot \tilde V ^{\top},
\end{equation}
Next, we leverage the InfoNCE loss \cite{oord2018representation} to pull semantically close pairs of  $\tilde E[W]_k$ and $\tilde V_k$ together and push away non-related pairs:
\begin{equation}
\label{infonce}
\begin{gathered}
\mathcal{L}_{t 2 v}=-\frac{1}{B} \sum_k^B \log \frac{e^{ \left(\text{sim}\left(\tilde E[W]_k, \tilde V_k\right) / \tau\right)}}{\sum_r^B e^{ \left(\text{sim}\left(\tilde E[W]_k, \tilde V_r\right) / \tau\right)}}, \\
\mathcal{L}_{v 2 t}=-\frac{1}{B} \sum_k^B \log \frac{e^{ \left(\text{sim}\left( \tilde V_k, \tilde E[W]_k\right) / \tau\right)}}{\sum_r^B e^{ \left(\text{sim}\left(\tilde V_k, \tilde E[W]_r\right) / \tau\right)}},\\
\mathcal{L}_{ccr}=\frac{1}{2}(\mathcal{L}_{t 2 v}+\mathcal{L}_{v 2 t}),
\end{gathered}
\end{equation}
where $\tau$  is the temperature hyper-parameter.  This loss formulates a self-supervisory signal to regularize the cross-modal alignment during caption generation, so as to improve the quality of generated captions.

\subsection{Joint Training}

We train the overall architecture  end-to-end by maximizing the likelihood of the observed  word sequence. Given the ground-truth words $\left(w_{1}^{*}, \ldots, w_{m}^{*}\right)$, we minimize the negative log-likelihood loss:
\begin{equation}
\mathcal L_{cap}(\theta)=-\sum_{t=1}^{m} \log p_\theta \left(w_{t}^{*} \mid w_{<t}^{*}\right),
\end{equation}
where $p_\theta\left(w_{t}^{*} \mid w_{<t}^{*}\right)$ is computed by Eq.~(\ref{word}), and $\theta$ are all the learnable parameters. In addition, our method is self-supervised by the losses of DIRL and CCR. Thus, the total loss   is defined as:
\begin{equation}
\label{cross-entropy}
\mathcal L =\mathcal L_{cap} + \lambda_d \mathcal{L}_{dirl} + \lambda_c \mathcal{L}_{ccr},
\end{equation}
where $\lambda_{d}$ and $\lambda_{c}$ are the trade-off parameters that are discussed in Sec. \ref{dirl_ccr}.

\section{Experiments}
\label{experiment}
\subsection{Datasets and Evaluation Metrics}
\textbf{Spot-the-Diff} \cite{jhamtani2018learning} has 13,192 image pairs from surveillance cameras, where each image pair includes an underlying illumination change.  According to the official split, we split it into training, validation, and testing with a ratio of 8:1:1.

\noindent \textbf{CLEVR-Change} \cite{park2019robust}  has 79,606 image pairs and 493,735 captions. It contains five change types, \emph{i.e.}, ``Color'', ``Texture'', ``Add'', ``Drop'',  and ``Move'', as well as moderate distractors. We use the official split of 67,660 for training, 3,976 for validation, and 7,970 for testing, respectively.

\noindent \textbf{CLEVR-DC} \cite{kim2021agnostic} has 48,000 image pairs with the same change types as CLEVR-Change, but this dataset includes drastic distractors. We use the official split with  85\% for training, 5\% for validation, and 10\% for testing. 

\noindent \textbf{Image Editing Request} \cite{tan2019expressing} is comprised of 3,939  image pairs with 5,695 editing  instructions.  We use the official split with 3,061 image pairs for training, 383 for validation, and 495 for testing, respectively.

\noindent \textbf{Evaluation Metrics:} We follow the state-of-the-art methods to use
the following five metrics for evaluating the quality of generated captions: BLEU-4  \cite{papineni2002bleu}, METEOR \cite{banerjee2005meteor}, ROUGE-L  \cite{lin2004rouge}, CIDEr \cite{vedantam2015cider} and SPICE \cite{anderson2016spice}.  We compute all the results by the Microsoft COCO evaluation server \cite{chen2015microsoft}.

\subsection{Implementation Details}
For fair-comparison, we follow current SOTA methods to utilize a pre-trained ResNet-101  \cite{he2016deep} to extract the features of a pair of images, with the dimension of 1024 $\times$ 14 $\times$ 14. We project them into a lower dimension of 512. We set the hidden size of the model and word embedding size as 512 and 300. Temperature $\tau$ in Eq. (\ref{infonce}) is set to 0.5. 
We train the model to converge with 10K iterations in total. Adam optimizer \cite{kingma2014adam} is employed to minimize the negative log-likelihood loss of Eq. (\ref{cross-entropy}). During inference, we use the greedy decoding strategy to generate captions. More details are shown in the supplementary material.

\begin{table}[]
  \centering

\caption{Comparison with SOTA methods on the Spot-the-Diff dataset. }
  \begin{tabular}{c|ccccc}
    \toprule
    Model & BLEU-4     & METEOR  & ROUGE-L   & CIDEr     & SPICE \\
    \midrule
    M-VAM \cite{shi2020finding} (ECCV 2020) & 10.1  & 12.4 & 31.3 & 38.1  & 14.0 \\
    DUDA+TIRG \cite{hosseinzadeh2021image} (CVPR 2021) & 8.1   & 12.5 & 29.9 & 34.5  & - \\
    R$^{3}$Net+SSP \cite{tu-etal-2021-r} (EMNLP 2021) & -     & 13.1 & \underline{32.6} & 36.6  & 18.8 \\
    VACC \cite{kim2021agnostic} (ICCV 2021) & 9.7   & 12.6 &32.1 & 41.5  & - \\
    MCCFormers-D \cite{qiu2021describing} (ICCV 2021) & 10.0  & 12.4 & - & \textbf{43.1} & 18.3 \\
    IFDC \cite{huang2022image} (TMM 2022) & 8.7   & 11.7 & 30.2 & 37.0  & - \\
    I3N-TD \cite{yue2023i3n} (TMM 2023) & \textbf{10.3}  & 13.0 & 31.5 & \underline{42.7}  & 18.6 \\
    VARD-Trans \cite{tu2023viewpoint} (TIP 2023) & -     & 12.5 & 29.3 & 30.3  & 17.3 \\
    SCORER+CBR \cite{tu2023self} (ICCV 2023) & \underline{10.2}  & 13.2  & - & 38.9 & 18.4 \\
    SMART \cite{tu2024smart} (TPAMI 2024) & - & \underline{13.5}    & 31.6 & 39.4 & \underline{19.0} \\
    \textbf{DIRL+CCR (Ours)} & \textbf{10.3} & \textbf{13.8} & \textbf{32.8} & 40.9  & \textbf{19.9} \\
    \bottomrule
    \end{tabular}%
    
  \label{com_spot}%
\end{table}%

\subsection{Performance Comparison}

\textbf{Results on the Spot-the-Diff Dataset.}
This dataset contains image pairs from  surveillance cameras with underlying illumination changes. We compare DIRL+CCR with the state-of-the-art methods: M-VAM \cite{shi2020finding}, DUDA+TIRG \cite{hosseinzadeh2021image}, R$^{3}$Net+SSP \cite{tu-etal-2021-r}, VACC \cite{kim2021agnostic},  MCCFormers-D \cite{qiu2021describing}, IFDC \cite{huang2022image}, I3N-TD \cite{yue2023i3n},  VARD-Trans \cite{tu2023viewpoint}, SCORER+CBR \cite{tu2023self}, and SMART \cite{tu2024smart}.

The results are shown in  Table \ref{com_spot}. Our DIRL+CCR obtains the best results on most metrics. Compared with the latest works SCORER+CBR and SMART, DIRL+CCR surpasses them on all metrics, especially with increases of 8.2\% and 4.7\% on SPICE. As the image pairs on this dataset are from surveillance cameras, most objects have no good postures, which makes the model difficult to locate and describe changed objects. In this situation, our DIRL+CCR also achieves an encouraging performance, which shows its robustness.
\begin{table}[t]
  \centering
 \caption{Comparison with SOTA methods on the CLEVR-Change dataset. }
    \begin{tabular}{c|ccccc}
    \toprule
    Model &  BLEU-4     & METEOR  & ROUGE-L   & CIDEr     & SPICE \\
    \midrule
    DUDA \cite{park2019robust} (ICCV 2019) & 42.9  & 29.7  & -     & 94.6  & 19.9 \\
    DUDA+TIRG \cite{hosseinzadeh2021image} (CVPR 2021) & 49.9  & 34.3  & 65.4  & 101.3 & 27.9 \\
    MCCFormers-D \cite{qiu2021describing} (CVPR 2021) & 53.3  & 37.1  & 70.8  & 119.1 & 30.4 \\
    R$^{3}$Net+SSP \cite{tu-etal-2021-r} (EMNLP 2021) & 52.7  & 36.2  & 69.8  & 116.6 & 30.3 \\
    IFDC \cite{huang2022image} (TMM 2022) & 47.2  & 29.3  & 63.7  & 105.4 & - \\
    I3N \cite{yue2023i3n} (TMM 2023) & 53.1  & 37.0  & 70.8  & 117.0     & 32.1 \\
    NCT \cite{tu2023neighborhood} (TMM 2023) & 53.1  & 36.5  & 70.7  & 118.4 & 30.9 \\
    VARD-Trans \cite{tu2023viewpoint} (TIP 2023) & 53.6  & 36.7  & 71.0  & 119.1 & 30.5 \\
    SCORER+CBR \cite{tu2023self} (ICCV 2023) & \underline{54.4}  & \underline{37.6}  & 71.7 & \underline{122.4} & 31.6 \\
    SMART \cite{tu2024smart} (TPAMI 2024) & 54.3  & 37.4  & \underline{71.8}  & \textbf{123.6} & \textbf{32.0} \\
    \textbf{DIRL+CCR (Ours) } & \textbf{54.6} & \textbf{38.1} & \textbf{71.9} & \textbf{123.6} & \underline{31.8} \\
    \bottomrule
    \end{tabular}%
   
  \label{com_change}%
\end{table}%

\begin{table}[t]
  \centering
    \caption{Evaluation on CLEVR-Change with varied change categories by METEOR.}
\begin{tabular}{c|ccccc}
    \toprule
          & \multicolumn{4}{c}{METEOR} \\
    \midrule
    Model & Color    & Texture     & Add    & Drop      & Move \\
    \midrule
    DUDA \cite{park2019robust} (ICCV 2019) & 32.8 & 27.3  & 33.4 & 31.4  & 23.5 \\
    DUDA+TIRG \cite{hosseinzadeh2021image} (CVPR 2021) & 36.1 & 30.4  & 37.8 & 36.7  & 27.0 \\
    R$^{3}$Net+SSP \cite{tu-etal-2021-r} (EMNLP 2021) & 38.9 & 35.5  & 38.0 & 37.5  & 30.9 \\
    IFDC \cite{huang2022image} (TMM 2022) & 33.1 & 27.9  & 36.2 & 31.4  & 31.2 \\
    I3N \cite{yue2023i3n} (TMM 2023) & 39.9 & 36.7  & \underline{39.9}  & \textbf{38.1} & 30.6 \\
    NCT \cite{tu2023neighborhood} (TMM 2023) & 39.1 & 36.3  & 39.0 & 37.2  & 30.5 \\
    SMART \cite{tu2024smart} (TPAMI 2024)  & \underline{40.2} & \underline{37.8}  & 39.3 &\textbf{38.1} & \underline{31.5} \\
    \textbf{DIRL+CCR (Ours) } & \textbf{40.7} & \textbf{38.2}  & \textbf{40.0} & \underline{37.9} & \textbf{33.5} \\
    \bottomrule
    \end{tabular}%
  
  \label{type_change}%
\end{table}%

\begin{table}[t]
  \centering
   \caption{Comparison with SOTA methods on the CLEVR-DC dataset.}
  \begin{tabular}{c|ccccc}
    \toprule
    Model & BLEU-4     & METEOR  & ROUGE-L   & CIDEr    & SPICE \\
    \midrule
    DUDA \cite{park2019robust} (ICCV 2019) & 40.3  & 27.1 & - & 56.7  & 16.1 \\
    M-VAM \cite{shi2020finding} (ECCV 2020) & 40.9  & 27.1 & - & 60.1  & 15.8 \\
    VACC \cite{kim2021agnostic} (ICCV 2021) & 45.0  & 29.3 & - & 71.7  & \textbf{17.6} \\
    NCT \cite{tu2023neighborhood} (TMM 2023) & 47.5  & \underline{32.5} & 65.1  & 76.9  & 15.6 \\
    VARD-Trans \cite{tu2023viewpoint} (TIP 2023) & 48.3  & 32.4 & - & 77.6  & 15.4 \\
    SCORER+CBR \cite{tu2023self} (ICCV 2023) & \underline{49.4}  & \textbf{ 33.4} & \underline{66.1} & \underline{83.7}  & 16.2 \\
    \textbf{DIRL+CCR (Ours)} & \textbf{51.4} & 32.3 & \textbf{66.3} & \textbf{84.1} & \underline{16.8} \\
    \bottomrule
    \end{tabular}%
  \label{com_dc}%
\end{table}%

\textbf{Results on the CLEVR-Change Dataset.}
This dataset contains basic geometric objects and moderate distractors.  We evaluate DIRL+CCR under the setting of semantic change \& distractors and detailed change categories. The comparison state-of-the-art methods are: DUDA \cite{park2019robust},  DUDA+TIRG \cite{hosseinzadeh2021image}, R$^{3}$Net+SSP \cite{tu-etal-2021-r},  MCCFormers-D \cite{qiu2021describing}, IFDC \cite{huang2022image}, I3N \cite{yue2023i3n},  VARD-Trans \cite{tu2023viewpoint}, SCORER+CBR \cite{tu2023self}, and SMART \cite{tu2024smart}. 

The results are shown in Table \ref{com_change}-\ref{type_change}.
In the both tables, our DIRL+CCR gains the superior results under the setting of semantic change \& distractors, and obtains best results on most change categories. The recent work NCT \cite{tu2023neighborhood} is a match-based method under the architecture of transformer, which directly correlates two image representations to extract the shared features and then removes them to compute the difference features. We find that DIRL+CCR outperforms it on all metrics. This comparison validates the effectiveness of performing distractors-immune representation learning before change localization, and cross-modal alignment regularizing during caption generation.

\textbf{Results on the CLEVR-DC Dataset.}
The experiment is conducted on CLEVR-DC with extreme distractors. We compare with the following state-of-the-art methods: DUDA \cite{park2019robust}, M-VAM \cite{shi2020finding}, VACC \cite{kim2021agnostic}, NCT \cite{tu2023neighborhood}, VARD-Trans \cite{tu2023viewpoint}, and SCORER+CBR \cite{tu2023self}. 
The results are shown in Table \ref{com_dc}. 

We note that the overall performance of our DIRL+CCR is better. Specifically,   DIRL+CCR obtains the best results on BLEU-4, ROUGE-L, and CIDEr metrics, while achieving comparable performance against the state-of-the-art methods on the other metrics. Especially, DIRL+CCR obtains relative improvement  of 8.2\%, 6.4\% and  4.0\% on BLEU-4 against the recent works NCT, VARD-Trans  and SCORER+CBR, respectively.  The results validate the robustness of our method under the disturbance of drastic distractors.

\begin{table}[t]
  \centering
  \caption{Comparison with the SOTA methods on  Image Editing Request. ``*'' indicates that the model is trained by pre-training and fine-tuning strategy.}
  \begin{tabular}{c|cccc}
    \toprule
    \multicolumn{1}{c|}{Model} & BLEU-4     & METEOR     & ROUGE-L     & CIDEr \\
   \midrule
    VIXEN-QFormer* \cite{black2024vixen} (AAAI 2024) & 7.9   & 14.4  & 33.5  & 35.4 \\
    VIXEN-CLIP* \cite{black2024vixen} (AAAI 2024) & 8.6   & \textbf{15.4}  & \textbf{42.5}  & \textbf{38.1} \\
    \midrule
    DUDA \cite{park2019robust} (ICCV 2019) & 6.5   & 12.4  & 37.3  & 22.8 \\
    Dyn rel-att \cite{tan2019expressing} (ACL 2019) & 6.7   & 12.8  & 37.5  & 26.4 \\
    BDLSCR \cite{yao2022image} (IJIS 2022) & 6.9   & 14.6  & 38.5  & 27.7 \\
    NCT \cite{tu2023neighborhood} (TMM 2023) & 8.1   & \underline{15.0} & 38.8  & 34.2 \\
    VARD-Trans \cite{tu2023viewpoint} (TIP 2023) & 10.0 & 14.8  & 39.0  & \underline{35.7} \\
    SCORER+CBR \cite{tu2023self} (ICCV 2023) & 10.0 & \underline{15.0}  & \underline{39.6}  & 33.4 \\
    SMART \cite{tu2024smart} (TPAMI 2024) & \underline{10.5} & \textbf{15.2}  & 39.1  & \textbf{37.8} \\
    \textbf{DIRL+CCR (Ours) }  & \textbf{10.9}   & \underline{15.0}  & \textbf{41.0} & 34.1 \\
    \bottomrule
    \end{tabular}%
    
  \label{com_edit}%
\end{table}%

\textbf{Results on the Image Editing Request Dataset.}
To further validate the generalization, we conduct the experiment on the scenario of image editing. The comparison methods are DUDA \cite{park2019robust}, Dynamic rel-att \cite{tan2019expressing},  BDLSCR \cite{yao2022image}, NCT \cite{tu2023neighborhood}, VARD-Trans \cite{tu2023viewpoint}, SCORER+CBR \cite{tu2023self}, and SMART \cite{tu2024smart}. Besides, we compare our method with the latest work VIXEN \cite{black2024vixen}. VIXEN is first pre-trained on a large-scale image editing dataset (which has not been released yet) and fine-tuned on the Image Editing Request dataset.

The results are shown in Table \ref{com_edit}. Compared with the end-to-end training methods, our DIRL+CCR achieves the best results on two metrics and is on par with the SOTA methods on other two metrics. Compared with VIXEN, our DIRL+CCR also achieves a competitive performance. The comparison results validate the effectiveness and generalization of our method. 

\textbf{Performance Analysis.} In brief, compared to the state-of-the-art methods, our method achieves superior results in different  scenarios. This benefits from that 1) DIRL can make the paired image representations non-perturbational and discriminative under distractors. This helps learn robust difference features for  caption generation.  2) CCR regularizes the cross-modal alignment by maximizing the  contrastive alignment between the features of generated words and attended difference, thus helping improve the quality of yielded captions.

\begin{table}[t]
  \centering
   \caption{Ablation study about DIRL and CCR on the CLEVR-DC dataset.}
    \begin{tabular}{c|c|c|cccc}
    \toprule
    Ablation & DIRL  & CCR  & BLEU-4     & ROUGE-L     & CIDEr     & SPICE \\
    \midrule
    Transformer &   $\times$    &  $\times$     & 48.9  & 65.6  & 79.6  & 15.7 \\
    Transformer &    $\checkmark$   &   $\times$    & 50.5  & 65.8  & 81.8  & 16.2 \\
    Transformer &     $\times$  &   $\checkmark$   & 49.3  & 65.5  & 82.7  & 16.4 \\
    Transformer &   $\checkmark$    &   $\checkmark$   & \textbf{51.4}  & \textbf{66.3}  & \textbf{84.1}  & \textbf{16.8} \\
    \bottomrule
    \end{tabular}%
    
  \label{aba_dc}%
\end{table}%

\begin{table}
  \centering
   \caption{Ablation study about DIRL on the CLEVR-DC dataset.}
    \begin{tabular}{c|c|c|cccc}
    \toprule
    Ablation & non-perturbation & discrimination & BLEU-4     & ROUGE-L     & CIDEr     & SPICE \\
    \midrule
    Transformer &   $\times$    &   $\times$    & 48.9  & 65.6  & 79.6  & 15.7 \\
    DIRL  &   $\checkmark$     &   $\times$    & \textbf{50.7} & 65.2  & 79.3  & \textbf{16.2} \\
    DIRL  &    $\times$   &     $\checkmark$   & 50.1  & 65.1  & 79.5  & \textbf{16.2} \\
    DIRL  &    $\checkmark$    &  $\checkmark$      & 50.5  & \textbf{65.8} & \textbf{81.8} & \textbf{16.2} \\
    \bottomrule
    \end{tabular}%
   
  \label{aba_DIRL}%
\end{table}%

\begin{figure}[t]
\centering
\includegraphics[width=1\textwidth]{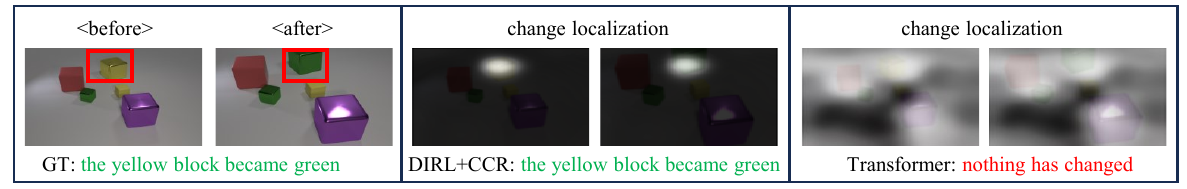} 
\caption{Visualization of change localization and captioning results of DIRL+CCR and Transformer. GT is short for ground-truth and changed objects are shown in  red boxes. }
\label{w/DIRL}
\end{figure}

\subsection{Ablation Study}
\textbf{Ablation study of each module.} To figure out the contribution of the proposed DIRL and CCR, we conduct the ablation study  on the CLEVR-DC dataset (with extreme distractors).  The baseline model is a vanilla Transformer. It directly interacts two image representations to extract share features, which are then removed from image representations to attain the difference features for caption generation.  
The results shown  in Table \ref{aba_dc}. 

When we augment the vanilla Transformer with DIRL, it yields a performance boost. This validates  DIRL can learn the two image representations that are non-perturbational and discriminative under the extreme distractors. As such, the model can better mine the shared features to learn reliable difference features.  Besides, the performance of vanilla Transformer is enhanced by equipping it with CCR, which verifies that CCR can regularize the cross-modal alignment. By using both DIRL and CCR, the vanilla Transformer achieves best performance, in particular with the improvements of 5.7\% and 7.0\% on CIDEr and SPICE metrics. This shows that each module not only plays its unique role, but also supplements each other.   In Fig. \ref{w/DIRL}, we  visualize  the  change localization and captioning results of DIRL+CCR and vanilla Transformer under distractors, where the unchanged objects appear obviously pseudo changes about scale and location. The visualization indicates that with the help of DIRL and CCR, the model can  pinpoint and describe the actually changed object.

\textbf{Ablation study of DIRL.} DIRL aims to learn two stable image representations from the perspectives of non-perturbation and discrimination, where the former is achieved by the first term of Eq. (\ref{alpha});  the latter is achieved by the second term. To validate our claim, we conduct the ablative study for non-perturbation and discrimination, respectively. The results are shown in Table \ref{aba_DIRL}.  The performance of DIRL with  either characteristic does not obtain much gains against the vanilla Transformer. Our conjecture is that the influence of distractors is not well solved. By jointly modeling the non-perturbation and discrimination, the performance of DIRL is significantly improved, which shows that each characteristic is key to handle distractors. That is, only if the model learns both, can it  make the paired image representations stable under distractors. As such, the model can capture the reliable difference features for caption generation.

\begin{figure}[t]
	\centering
	\begin{minipage}{0.49\linewidth}
		\centering
		\includegraphics[width=0.9\linewidth]{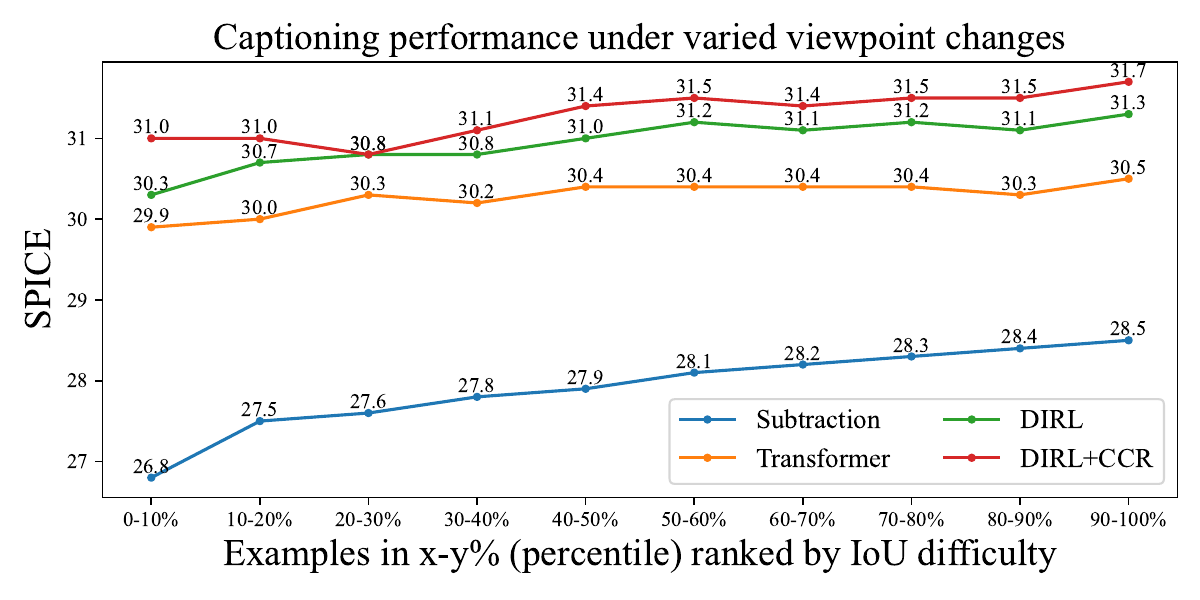}
		\caption{Visualization of captioning performance  under varied viewpoint changes.}
		\label{iou}
	\end{minipage}
 \hspace{0.05pt}
        \begin{minipage}{0.49\linewidth}
		\centering
		\includegraphics[width=0.9\linewidth]{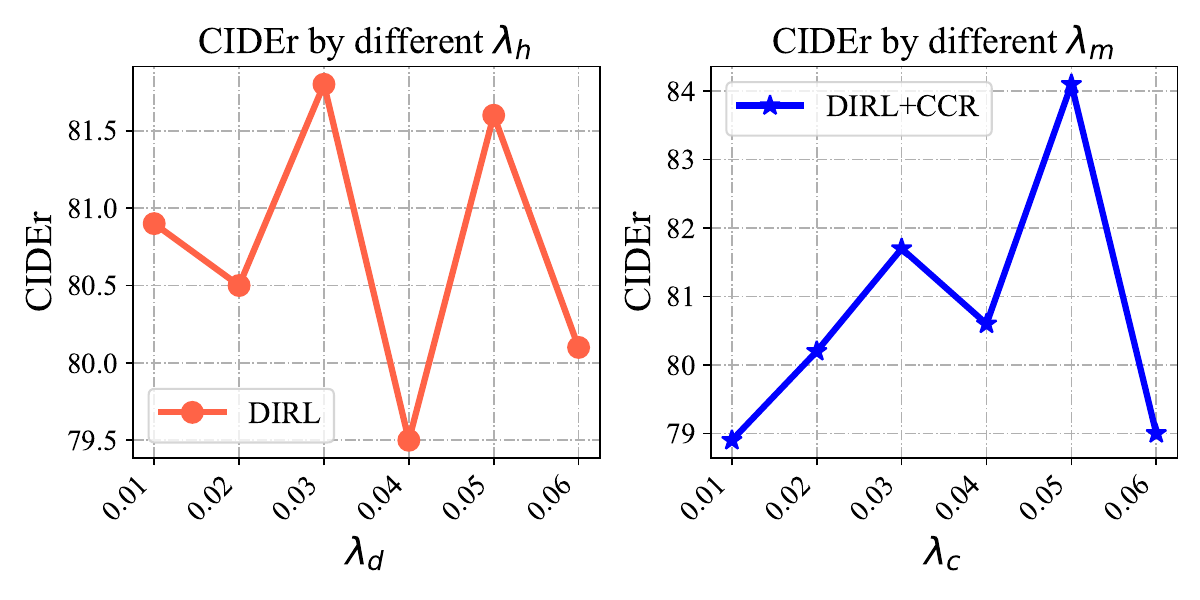}
		\caption{The effects of two trade-off parameters of $\lambda_d$ and
 $\lambda_c$ on  CLEVR-DC.}
		\label{cider}
	\end{minipage}
\end{figure}

\subsection{Measuring Captioning Performance under Distractors}
To verify whether our method is immune to distractors, we measure the model's captioning performance under varied viewpoint changes. The varied viewpoints are computed by the IoU of the bounding boxes of the objects of two images, where the more the camera changes its position, the less the bounding boxes overlap. That is, 
lower IoU means higher difficulty. The results are shown in Fig.  \ref{iou}, where the compared models are Subtraction (direct subtraction between two image representations), vanilla Transformer, DIRL and DIRL+CCR.  We find that the performance of Subtraction is the lowest, showing that direct subtraction generalizes poorly to distractors.  DIRL apparently surpasses Transformer, validating it does make the representations of image pair immune to distractors. Further, the performance of DIRL further boosts by augmenting it with CCR, which demonstrates that CCR plays a role in enhancing cross-modal alignment and thus improves the captioning quality  under distractors.

\subsection{Study on the Trade-off Parameters $\lambda_d$ and $\lambda_c$}
\label{dirl_ccr}
We discuss the effects of two trade-off parameters $\lambda_d$ and $\lambda_c$ in Eq.(\ref{cross-entropy}). Fig. \ref{cider} shows the results  of DIRL and DIRL+CCR with different  $\lambda_d$ and $\lambda_c$ on  the CLEVR-DC dataset. We first discuss  $\lambda_d$. As its value  increases or decreases, the performance of DIRL changes. Based the results, we set $\lambda_d$ as 0.03. Then, we fix $\lambda_d$ to discuss  $\lambda_c$ and set it as  $0.05$. Similarly, we further discuss $\lambda_d$ and $\lambda_c$ on the other three datasets, and set them as  0.5 and 0.3 on the CLEVR-Change dataset; 0.5 and 0.004 on the Spot-the-Diff dataset; 0.001 and 0.05 on the Image Editing Request dataset.

\begin{figure*}[t]
\centering
\includegraphics[width=1\textwidth]{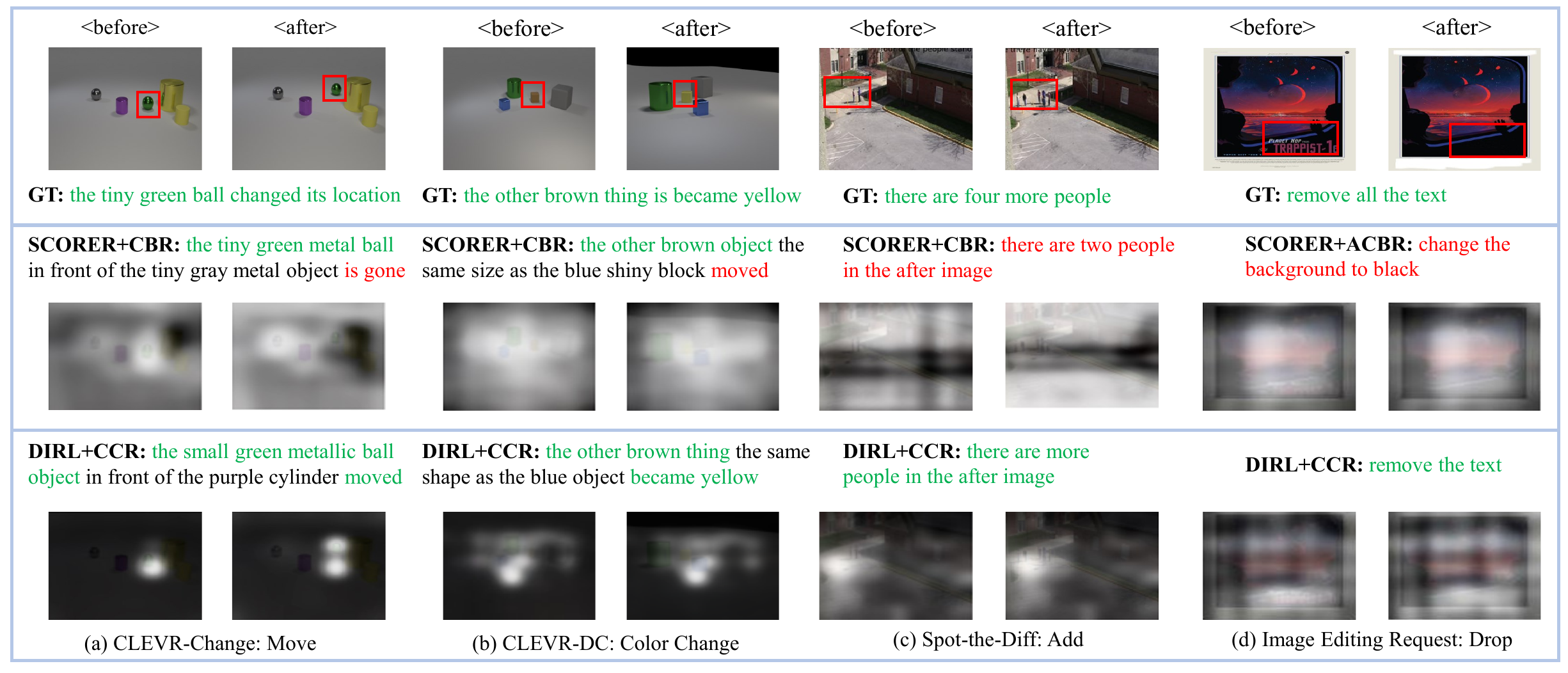} 
\caption{Qualitative analysis on four datasets. We visualize the ground-truth captions (GT), generated captions and change localization results of SCORER+CBR \cite{tu2023self} and our DIRL+CCR. The ground-truth changes are shown in  red boxes.}  
\label{example}
\end{figure*}

\subsection{Qualitative Analysis}

We conduct qualitative analysis about change localization and caption generation on the four datasets, as shown in Fig. \ref{example} (a)-(d). The compared method is SCORER+CBR \cite{tu2023self}, which first obtains view-invariant representations  and then interacts both to compute the difference features, as well as improving captioning quality by cross-model backward reasoning.
In these examples, the change localization maps of SCORER+CBR are relatively scattered and cannot focus on genuine changes. Hence, it fails to generate the correct sentences to describe these changes. By contrast, our DIRL+CCR successfully recognizes and captions the changes. This superiority benefits from  making two image representations non-perturbational and discriminative under distractors. As such, our method can better pinpoint the actually changed object between them. During caption generation, CCR  further regularizes cross-modal alignment, so as to make the target words generated based on the most related difference features. More qualitative examples are shown in the supplementary material.

\section{Conclusion}
\label{conclusion}
In this paper, we propose the DIRL to learn  a pair of non-perturbational and discriminative image representations under distractors, by correlating their corresponding channels and decorrelating different channels. As such, the model can sufficiently mine their shared features to learn the robust difference features for caption generation.  Further, we  design the CCR to regularize the cross-modal alignment by   maximizing the contrastive alignment between the attended difference features and generated words, which helps improve captioning performance.  Extensive experiments show that our method yields state-of-the-art results on the four public  datasets with different change scenes. 


\section*{Acknowledgements}
This work was supported in part by National Natural Science Foundation of China: 62322211, 61931008, 62236008, 62336008, U21B2038, 62225207, Fundamental Research Funds for the Central Universities (E2ET1104), ``Pionee'' and ``Leading Goose'' R\&D Program of Zhejiang Province (2024C01023).

\bibliographystyle{splncs04}
\bibliography{main}

\section{Appendix}
\subsection{Experiments}
\subsubsection{Implementation Details}
For a fair comparison, we follow the state-of-the-art methods \cite{tu2023self,tu2024smart} to utilize a pre-trained ResNet-101  \cite{he2016deep} model to obtain the features of a pair of images, with the dimension of 1024 $\times$ 14 $\times$ 14. We project them into a lower dimension of 512. The hidden size of the model and word embedding size are set to 512 and 300, separately. Temperature $\tau$ in Eq. (14) of main paper is set to 0.5. 

In the training phase, the batch sizes and learning rates of our method on the four datasets are shown in Table \ref{batch}.  We use Adam optimizer \cite{kingma2014adam}  to minimize the negative log-likelihood loss of Eq. (16) of main paper. In the inference phase, we use the greedy decoding strategy to generate captions. Both training and inference are implemented with PyTorch on an RTX 3090 GPU. The used training resources on the all datasets are shown in Table \ref{time}. We find that our method does not require much training time and GPU memory, so it can be easily reproduced by the researchers.  To facilitate future research, the code is publicly available at \url{https://github.com/tuyunbin/DIRL}.

\begin{table}[h]
			\centering
			 \caption{The training parameters of our method on the four datasets.}
    \begin{tabular}{c|c|c}
   \hline
          & batch size & learning rate \\
    \hline
    CLEVR-Change & 128  & 2 $\times$ $10^{-4}$ \\
    CLEVR-DC & 128 &  2 $\times$ $10^{-4}$ \\
    Spot-the-Diff & 64 &  1 $\times$ $10^{-4}$ \\
    Image Editing Request & 16 &  1 $\times$ $10^{-4}$ \\
    \hline
    \end{tabular}%
   
\label{batch}%
	\end{table} 

 \begin{table}[h]
			\centering
		 \caption{The used training time and GPU memory on the four datasets.}	
    \begin{tabular}{c|c|c}
    \hline
          & Training Time & GPU Memory \\
    \hline
    CLEVR-Change & 150 minutes  & 14 GB  \\
    CLEVR-DC & 90 minutes  & 8.4 GB \\
    Spot-the-Diff & 25 minutes  & 6.5 GB \\
    Image Editing Request & 10 minutes  & 3.2 GB \\
    \hline
    \end{tabular}%
   
  \label{time}%
		\end{table}

\begin{table}[h]
			\centering
		 \caption{Effect of trade-off parameter $\alpha$ in DIRL on the CLEVR-DC dataset.}	
    \begin{tabular}{c|ccccc}
    \toprule
       $\alpha$   & BLEU-4     & METEOR     & ROUGE-L     & CIDEr     & SPICE \\
       \midrule
    0.001 & \textbf{50.6}  & 32.3  & 65.7  & \textbf{81.8}  & 16.1 \\
    0.002 & 50.4  & \textbf{32.5}  & \textbf{65.9}  & 81.2  & 16.1 \\
    0.003 & 50.5  & \textbf{32.5}  & 65.8  & \textbf{81.8}  & 16.2 \\
    0.004 & 50.4  & 32.3  & 65.7  & 80.5  & \textbf{16.3} \\
    0.005 & 50.1  & 31.9  & 65.4  & 79.7  & 16.2 \\
    0.006 & 50.5  & 31.5  & 64.9  & 79.8  & 15.8 \\
    \bottomrule
    \end{tabular}%
   
  \label{alpha}%
		\end{table} 

\subsubsection{Study on the Trade-off Parameter $\alpha$}
In this section, we discuss the effect of trade-off parameter $\alpha$ in Eq. (4) of main paper. As mentioned in Sec. 3.2 Distractors-Immune Representation Learning, $\alpha$ is a trade-off parameter to balance the importance between the two terms.  The first term  equates the diagonal of $\mathcal{C}$ to one, \textit{i.e.,} the corresponding feature channels of two image representations will be correlated and thus have similar semantics under distractors.
 The second term equates the off diagonal of $\mathcal{C}$ to zero, \textit{i.e.,} the different channels will be decorrelated. This enhances the discrimination of each image representation. 
Both terms are key to handle the influence of distractors. Since the CLEVR-DC dataset is with extreme distractors and more challenging, we conduct experiment on this dataset. In Table \ref{alpha}, we find that the captioning results are close when setting the value of $\alpha$ from 0.001 to 0.006, and the  model's overall performance is better under the value of 0.003. This shows that the proposed DIRL is robust. Empirically, we set the value of  $\alpha$ as 0.003 on the four datasets.

\begin{table}[h]
    \centering
     \caption{Ablation of DIRL with/without the MLP on the CLEVR-DC dataset. }
    \begin{tabular}{c|ccccc}
    \toprule
    Ablation & BLEU-4     & METEOR     & ROUGE-L   & CIDEr    & SPICE \\
    \midrule
    DIRL w/ MLP & \textbf{51.4} & \textbf{32.3} & \textbf{66.3}  & \textbf{84.1} & \textbf{16.8} \\
    DIRL w/o MLP & 48.0  & 32.0  & 65.2   & 78.5  & 15.7 \\
    \bottomrule
    \end{tabular}%
   
    \label{mlp}
\end{table}

\subsubsection{Study on the MLP Function in DIRL}
In Eq.(1) of the main paper, a convolution function transforms the features of two images into a low dimension, and then the trainable position encodings are integrated into the transformed features along their height and width. Further, the MLP in Eq.(2) projects  the two position-embedded features into a shared embedding space. To study the effect of the MLP, we carry out the ablation study of DIRL with/without  MLP on the CLEVR-DC dataset, which is show in Table \ref{mlp}. It is noted that DIRL with MLP is much better than it without MLP, which shows that adding the MLP helps  DIRL learn a pair of distractors-immune  representations in terms of semantics and position.

\subsubsection{Comparison between DIRL and Static Methods}
The proposed DIRL aims to learn a pair of distractors-immune  representations by correlating the corresponding feature channels and decorrelating different ones. In fact, there are some simpler static methods such as PCA or ZCA whitening to remove the degree of correlation. Here, we conduct the experiment to show performance comparison of cross-channel decorrelation among the transformer-based model with PCA/ZCA whitening and our DIRL. 

\begin{table}[htbp]
  \centering
  \small
  \caption{Performance comparison among Transformer-based model with PCA/ZCA whitening and our DIRL. }
    \begin{tabular}{c|cccc}
    \toprule
    Ablative Variants & BLEU-4 & ROUGE-L & CIDEr & SPICE \\
    \midrule
    Transformer & 48.9  & 65.6  & 79.6  & 15.7 \\
    Transformer w/ PCA whitening & 40.6  & 57.9  & 22.0  & 10.0 \\
    Transformer w/ ZCA whitening & 38.6  & 57.3  & 33.3  & 13.1 \\
    Transformer w/ DIRL & \textbf{50.5}  & \textbf{65.8}  & \textbf{81.8}  & \textbf{16.2} \\
    \bottomrule
    \end{tabular}%
  \label{DIRL vs}%
\end{table}%

The two static methods are used as data preprocessing strategies before model training, while the proposed DIRL is jointly trained with the model. The comparison results are shown in Table \ref{DIRL vs}. We can find that the model with DIRL outperforms the others by a large margin, indicating that joint training strategy (DIRL) does help the model learn two stable image features under distractors.

\begin{figure*}[t]
\centering
\includegraphics[width=1\textwidth]{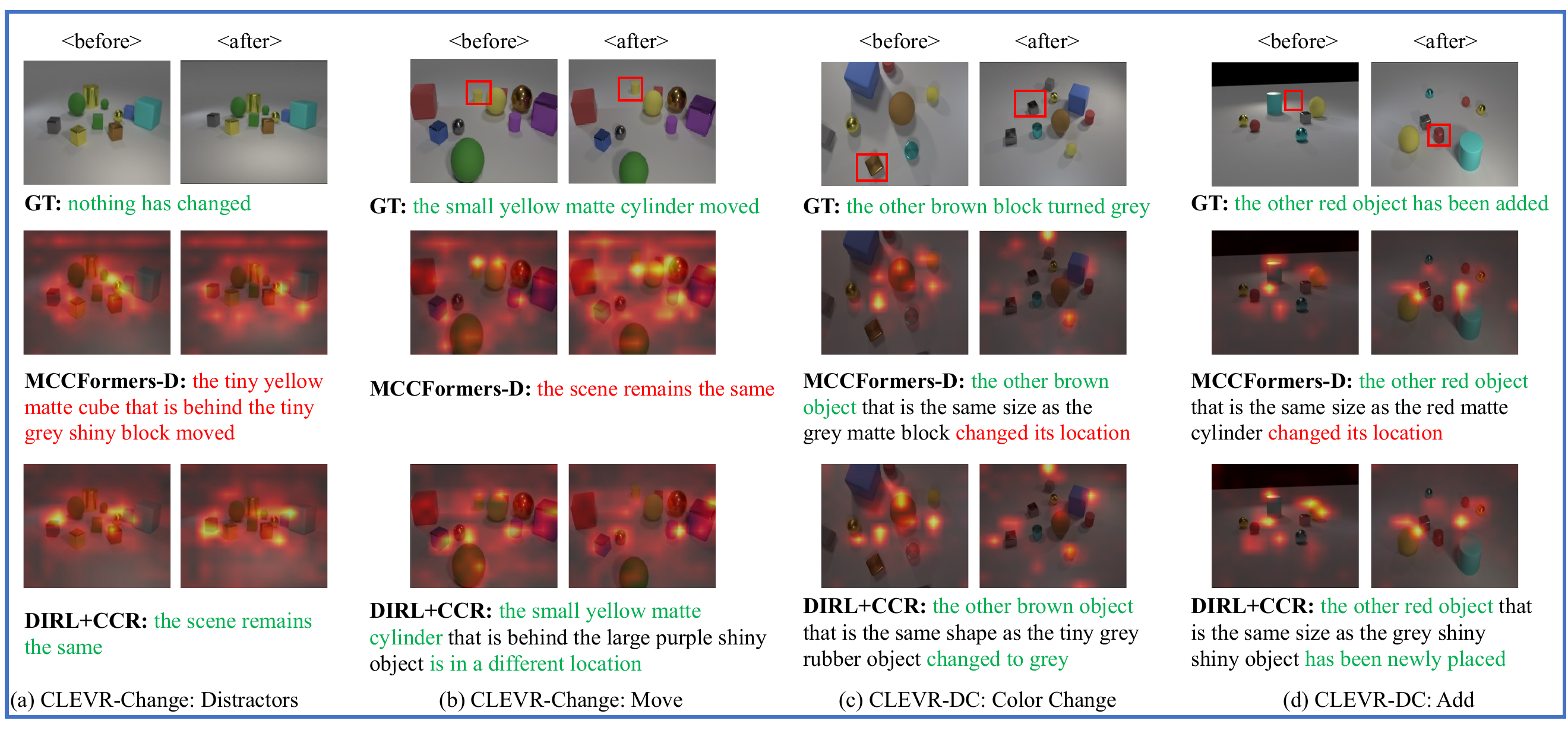} 
\caption{Visualization of shared objects matching between two images on the CLEVR-Change and CLEVR-DC with distractors to varying degrees. For each example, we  visualize matching results by the classic match-based method MCCFormers-D \cite{qiu2021describing} and the proposed DIRL+CCR.  The changed objects are shown in the red boxes.   Besides, we  visualize the ground-truth caption (GT), and the generated captions by  MCCFormers-D and our DIRL+CCR. The correct words are in green color, while incorrect words are in red color.  }
\label{dc_align}
\end{figure*}

\begin{figure*}[t]
\centering
\includegraphics[width=1\textwidth]{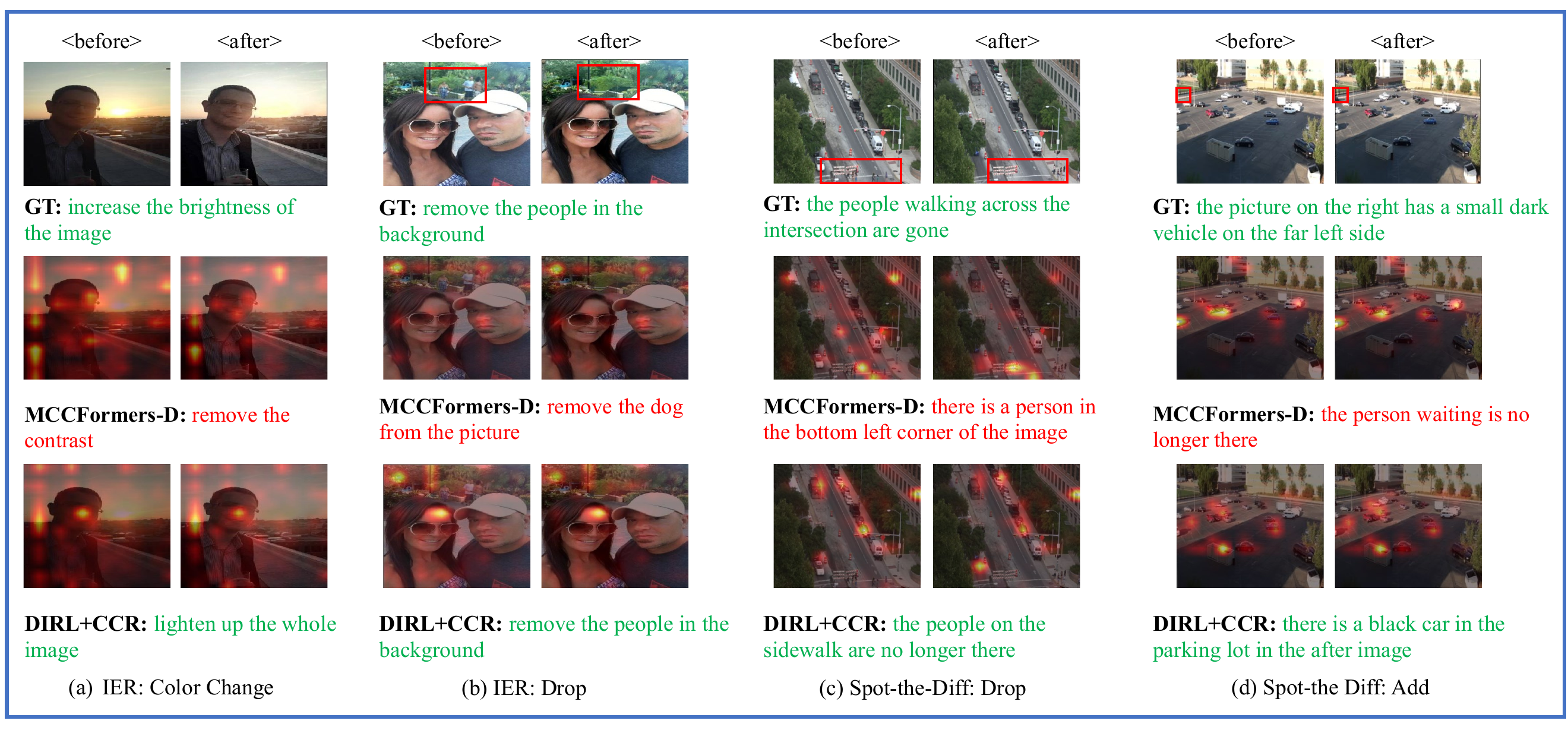} 
\caption{Visualization of shared objects matching between two images on the Image Editing Request (IER) and Spot-the-Diff datasets. For each example, we  visualize matching results by the classic match-based method MCCFormers-D \cite{qiu2021describing} and the proposed DIRL+CCR.  The changed objects are shown in the red boxes.   Besides, we  visualize the ground-truth caption (GT), and the generated captions by  MCCFormers-D and our DIRL+CCR. The correct words are in green color, while incorrect words are in red color.  }
\label{spot_align}
\end{figure*}

\subsubsection{Qualitative Analysis}
\begin{figure*}[h]
\centering
\includegraphics[width=1\textwidth]{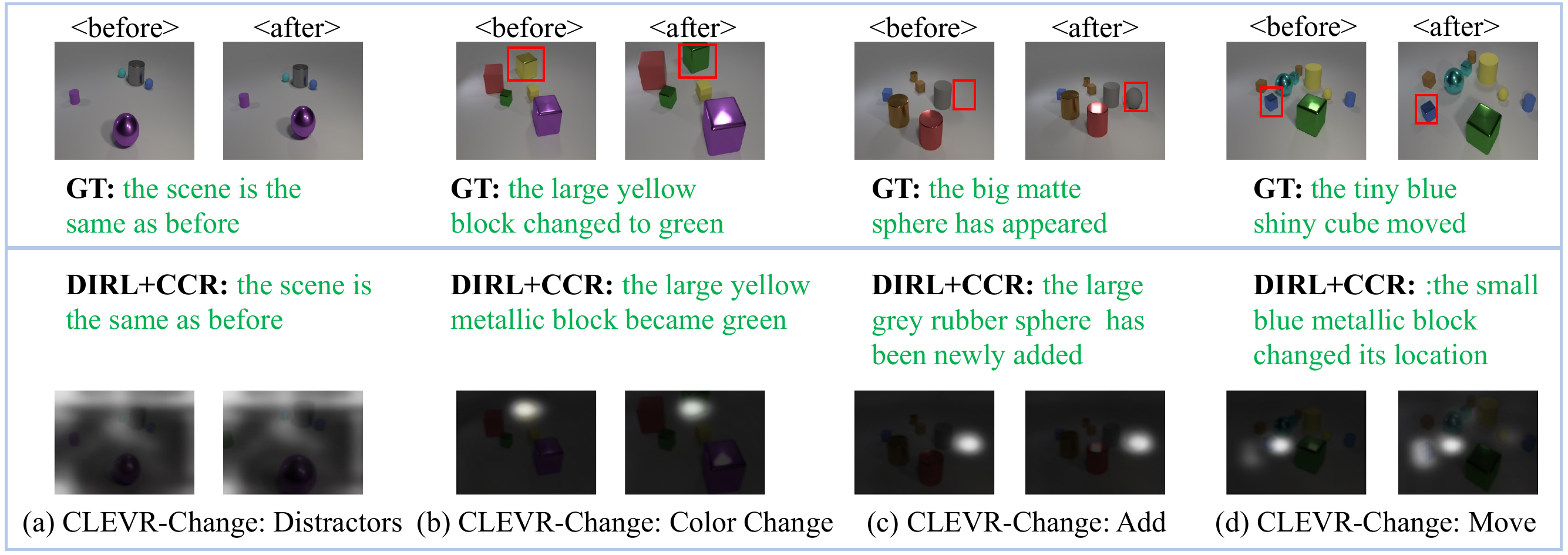} 
\caption{Four cases  from the CLEVR-Change dataset. We visualize the ground-truth caption (GT), and the captions yielded by our DIRL+CCR. We also visualize its change localization results. The ground-truth changed objects are shown in red boxes.}
\label{change_text}
\end{figure*}

\begin{figure*}[h]
\centering
\includegraphics[width=1\textwidth]{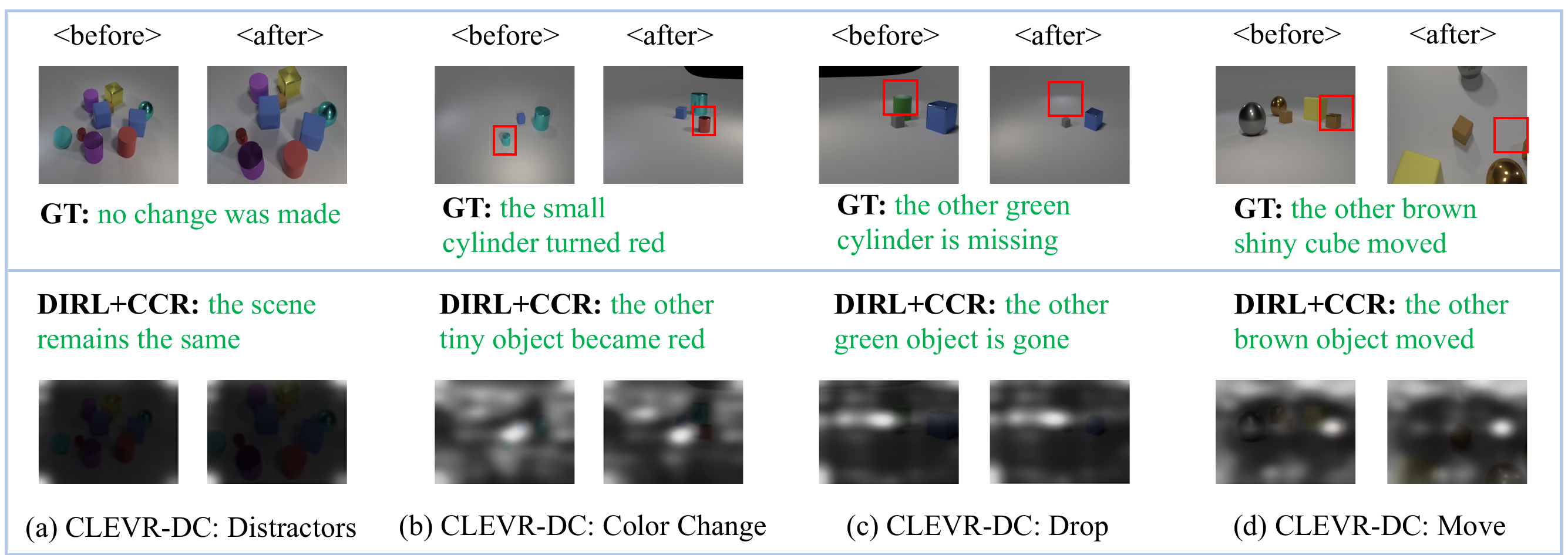} 
\caption{Four cases  from the CLEVR-DC dataset. We visualize the ground-truth caption (GT), and the captions yielded by our DIRL+CCR. We also visualize its change localization results. The ground-truth changed objects are shown in red boxes.}
\label{dc_text}
\end{figure*}

\begin{figure*}[t]
\centering
\includegraphics[width=1\textwidth]{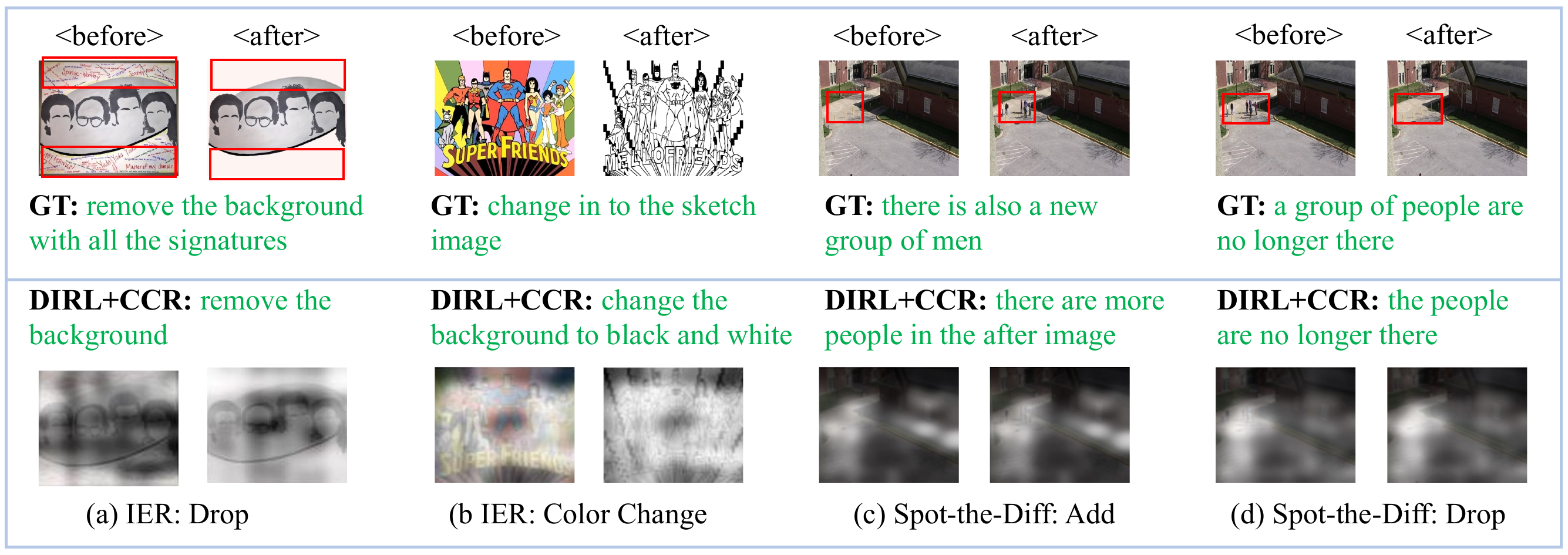} 
\caption{Four cases  from the Image Editing Request (IER) and Spot-the-Diff datasets. We visualize the ground-truth caption (GT), and the captions yielded by our DIRL+CCR. We also visualize its change localization results. The ground-truth changed objects are shown in red boxes.}
\label{spot_text}
\end{figure*}

In this supplementary material, we will show more qualitative examples on the CLEVR-Change, CLEVR-DC, Spot-the-Diff, and Image Editing Request datasets, which are shown in Fig. \ref{dc_align}-\ref{spot_text}. In Fig. \ref{dc_align}-\ref{spot_align}, on the four datasets, we visualize shared objects matching between two images, which are yielded by the classic match-based method MCCFormers-D  \cite{qiu2021describing} and the proposed DIRL+CCR. We can find that MCCFormers-D is unable to sufficiently or correctly align the shared objects between two images. By contrast, the proposed DIRL+CCR can better match the shared objects. These examples validate that the proposed DIRL can make the representations of image pair non-perturbational and discriminative under the distractors to varying degrees. Based on the two distractors-immune image representations, the model can better interact and mine their shared features.

To intuitively validate the change localization and caption capabilities of our method, we  visualize the generated captions by DIRL+CCR under different change types on the four datasets. Meanwhile, we visualize the change localization results that are obtained from the cross-attention maps between the difference features and generated words in the decoding process. These are shown in Fig. \ref{change_text}-\ref{spot_text}. When the attention score is higher, the region is brighter. We observe that the proposed DIRL+CCR can accurately localize and describe the actually changed objects under different scenarios. This superiority mainly benefits from the facts that 1) DIRL is able to learn two distractors-immune image representations for matching the shared objects, so as to learn the reliable difference features for caption generation; 2) CCR is capable of regularizing the cross-modal alignment by maximizing the contrastive alignment between the generated words and  attended difference features, so as to improve the quality of yielded captions.

\subsection*{Limitation} 

\begin{figure*}[t]
\centering
\includegraphics[width=0.6\textwidth]{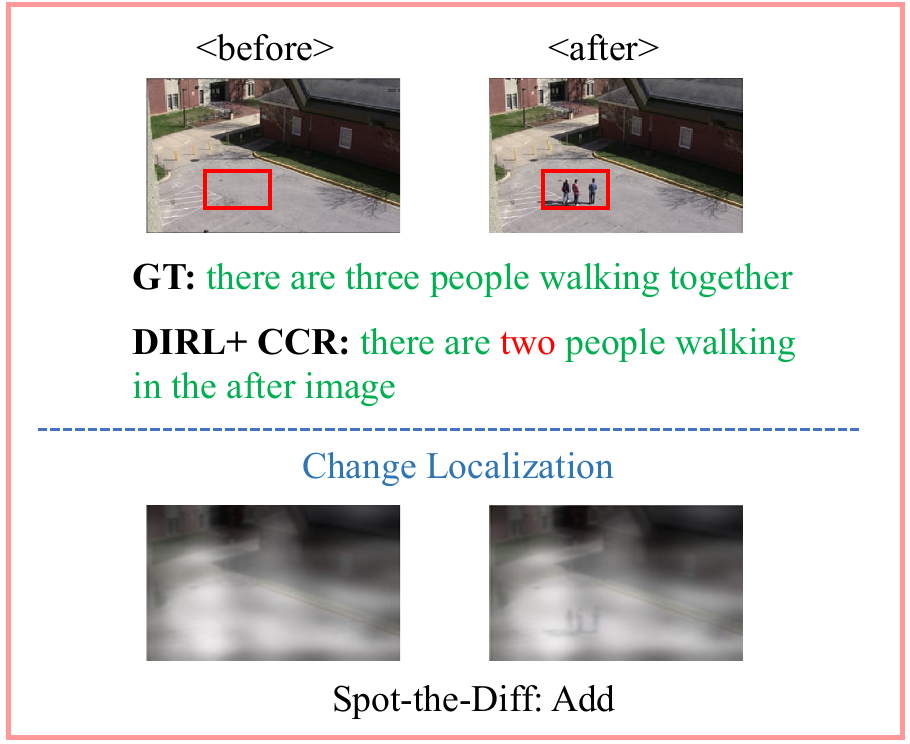} 
\caption{The failure case that derives from the proposed DIRL+CCR. }
\label{failure}
\end{figure*}

Fig. \ref{failure} shows a failure case that derives from our DIRL+CCR. This image pair is from the surveillance cameras and with underlying distractors (illumination change), where there are three people newly appearing in the ``after'' image. DIRL+CCR can accurately  localize the region containing the added people and describe the change type, which benefit from the proposed DIRL and CCR. 
However, we note that the generated sentence wrongly describes the number of added people. Our conjecture is that under surveillance cameras, the changed objects are commonly small. Further, the distance of the left two people is very close, and the color of one man's coat is similar to the other's pants. In this situation, the model risks recognizing two people as one person, so as to generate the incorrect result. In our opinion, a possible solution is to leverage finer-level visual modality for the representation of such small objects, \emph{e.g.,} image segmentation features \cite{zhang2020causal,kirillov2023segment}. 


\end{document}